\documentclass{article}
\pdfoutput=1
\PassOptionsToPackage{square,sort,comma,numbers}{natbib} %
\usepackage[final]{nips_2017}

\usepackage[utf8]{inputenc} %
\usepackage[T1]{fontenc}    %
\usepackage{url}            %
\usepackage{booktabs}       %
\usepackage{amsfonts}       %
\usepackage{nicefrac}       %
\usepackage{microtype}      %

\usepackage{mathtools}
\usepackage{bm}
\usepackage{dsfont}
\usepackage{color}
\usepackage{graphicx} %
\usepackage{subfigure}
\graphicspath{{final_figures/}, }
\usepackage{enumitem}

\newcommand{\mysubsection}[1]{\textbf{#1}:}
\newcommand{\evalsubsection}[1]{\textbf{#1}:}

\newcommand{\tdist}{p^\ast\!} 
\newcommand{\tdistx}{\tdist(\vx)}

\newcommand{\zdist}{p(\vz)}
\newcommand{\modeldist}{p_{\vtheta}}
\newcommand{\modeldistx}{p_{\vtheta}(\vx)}

\newcommand{\rphix}{r_{\vphi}(\vx)}

\newcommand{\vardistz}{q_{\eta}(\vz|\vx)}

\newcommand\cut[1]{}

\newcommand{\Reals}{\mathbb{R}}

\newcommand{\squishlist}{
   \begin{list}{$\bullet$}
    { \setlength{\itemsep}{0pt}      \setlength{\parsep}{3pt}
      \setlength{\topsep}{3pt}       \setlength{\partopsep}{0pt}
      \setlength{\leftmargin}{1.5em} \setlength{\labelwidth}{1em}
      \setlength{\labelsep}{0.5em} } }

\newcommand{\squishlisttwo}{
   \begin{list}{$\bullet$}
    { \setlength{\itemsep}{0pt}    \setlength{\parsep}{0pt}
      \setlength{\topsep}{0pt}     \setlength{\partopsep}{0pt}
      \setlength{\leftmargin}{2em} \setlength{\labelwidth}{1.5em}
      \setlength{\labelsep}{0.5em} } }

\newcommand{\squishend}{
    \end{list}  }

\newcommand{\KLpq}[2]{\textrm{KL}\!\left[{#1}||{#2}\right]}

\newcommand{\myvec}[1]{\mathbf{#1}}
\newcommand{\myvecsym}[1]{\boldsymbol{#1}}

\newcommand{\veta}{\myvecsym{\eta}}

\newcommand{\vomega}{\myvecsym{\omega}}

\newcommand{\vphi}{\myvecsym{\phi}}

\newcommand{\vtheta}{\myvecsym{\theta}}

\newcommand{\vx}{\myvec{x}}
\newcommand{\vxhat}{\hat{\vx}}

\newcommand{\vz}{\myvec{z}}
\newcommand{\vzhat}{\hat{\vz}}

\newcommand{\vI}{\myvec{I}}

\newcommand{\E}{\mathbb{E}}

\newcommand{\be}{\begin{equation}}
\newcommand{\ee}{\end{equation}}
\newcommand{\bea}{\begin{eqnarray}}
\newcommand{\eea}{\end{eqnarray}}
\newcommand{\beaa}{\begin{eqnarray*}}
\newcommand{\eeaa}{\end{eqnarray*}}

\DeclareMathAlphabet{\mathpzc}{OT1}{pzc}{m}{n}

\newcommand{\N}{\mathcal{N}}
\newcommand{\disc}{\mathcal{D}_{\vphi}}
\newcommand{\discx}{\mathcal{D}_{\vphi}(\vx)}
\newcommand{\codedisc}{\mathcal{C}_{\vomega}}
\newcommand{\codediscz}{\codedisc(\vz)}

\newcommand{\generator}{\mathcal{G}_{\vtheta}}

\usepackage{afterpage}
\usepackage{algorithm}
\usepackage{algorithmicx}
\usepackage{algcompatible}
\usepackage{algpseudocode}
\algrenewcommand\algorithmicindent{0.5em} %
\newcommand{\algcomment}[1]{\Comment{\textit{#1}}}
\algnewcommand{\LineComment}[1]{\(\triangleright\) \textit{#1}}
\newcommand{\addhspace}{\quad}
\newcommand{\addhspacesmall}{\hspace{1pt}}

\newcommand{\ourgan}{$\alpha$-GAN}
\newcommand{\colormnist}{ColorMNIST}

\definecolor{mydarkblue}{rgb}{0,0.08,0.45}
\usepackage[colorlinks,citecolor=mydarkblue,urlcolor=mydarkblue,linkcolor=mydarkblue,filecolor=mydarkblue]{hyperref}
\title{Variational Approaches for Auto-Encoding \\ Generative Adversarial Networks}%

\author{
Mihaela Rosca\thanks{Equal contribution.} \quad Balaji Lakshminarayanan\footnotemark[1] \quad
David Warde-Farley \quad Shakir Mohamed\\
  DeepMind\\
  \texttt{\{mihaelacr,balajiln,dwf,shakir\}@google.com}
}

\begin{document}

\maketitle

\begin{abstract}
\vspace{-2mm}
Auto-encoding generative adversarial networks (GANs) combine the standard GAN algorithm, which discriminates between real and model-generated data, with a reconstruction loss given by an auto-encoder. Such models aim to prevent mode collapse in the learned generative model by ensuring that it is grounded in all the available training data.
In this paper, we %
develop a principle upon which auto-encoders can be combined with generative adversarial networks by exploiting the hierarchical structure of the generative model.
The underlying principle shows that variational inference can be used a basic tool for learning, but with the intractable likelihood replaced by a synthetic likelihood, and the unknown posterior distribution replaced by an implicit distribution; both synthetic likelihoods and implicit posterior distributions can be learned using discriminators. This allows us to develop a natural fusion of variational auto-encoders and generative adversarial networks, combining the best of both these methods. We describe a unified objective for optimization, discuss the constraints needed to guide learning, connect to the wide range of existing work, and use a battery of tests to systematically and quantitatively assess the performance of our method.
\end{abstract}
\vspace{-4.5mm}

\section{Introduction}
\label{sect:intro}
\vspace{-3mm}
Generative adversarial networks (GANs) \citep{gan} are one of the dominant approaches for learning generative models in contemporary machine learning research, which provide a flexible algorithm for learning in latent variable models. Directed latent variable models describe a data generating process in which a source of noise is transformed into a plausible data sample using a non-linear function, and GANs drive learning by discriminating observed data from model-generated data. GANs allow for training on large datasets, are fast to simulate from, and when trained on image data, produce %
 visually compelling sample images. But this flexibility comes with instabilities in optimization that leads to the problem of mode-collapse, in which generated data does not reflect the diversity of the underlying data distribution. A large class of GAN variants that aim to address this problem are auto-encoder-based GANs (AE-GANs), that use an auto-encoder to
encourage the model to better represent \textit{all} the data it is trained with, thus discouraging mode-collapse.

Auto-encoders have been successfully used to improve GAN training. For example, plug and play generative networks (PPGNs) \citep{ppgn} produce state-of-the-art samples by optimizing an objective that combines an auto-encoder loss, a GAN loss, and a classification loss defined using a pre-trained classifier.
AE-GANs can be broadly classified into three approaches: (1) those using an auto-encoder as the discriminator, such as energy-based GANs and  boundary-equilibrium GANs \citep{began}, (2) those using a denoising auto-encoder to derive an auxiliary loss for the generator, such as denoising
feature matching GANs \citep{davidgan}, %
and (3) those combining ideas from VAEs and GANs. %
For example, the variational auto-encoder GAN (VAE-GAN) \citep{aepixels} adds  an adversarial loss to the  variational evidence lower bound objective.
More recent GAN variants, such as mode-regularized GANs (MRGAN) \citep{moderegularizedgan} and adversarial generator encoders (AGE) \citep{age} also use a separate encoder in order to stabilize GAN training.
Such variants are interesting because they reveal interesting connections to VAEs, however
the principles underlying the fusion of auto-encoders and GANs remain unclear. %

In this paper, we develop a principled approach for hybrid AE-GANs. By exploiting the hierarchical structure of the latent variable model learned by GANs, we show how another popular approach for learning latent variable models, variational auto-encoders (VAEs), can be combined with GANs. This approach will be advantageous since it allows us to overcome the limitations of each of these methods. Whereas VAEs often produce blurry images when trained on images, they do not suffer from the problem of mode collapse experienced by GANs.  GANs allow few distributional assumptions to be made about the model, whereas VAEs allow for inference of the latent variables which is useful for representation learning, visualization and explanation. The approach we will develop will combine the best of these two worlds, provide a unified objective for learning, is purely unsupervised, requires no pre-training or external classifiers, and can easily be extended to other generative modeling tasks.

We begin by exposing the tools that we acquire for dealing with intractable generative models from both GANs and VAEs in section \ref{sect:intractableModels}, and then make the following contributions:
\vspace{-1mm}
\begin{itemize}[leftmargin=3ex,topsep=0pt,itemsep=-1ex,partopsep=1ex,parsep=1ex]
\item We show that variational inference applies equally well to GANs and how discriminators can be used for variational inference with implicit posterior approximations.
\item Likelihood-based and likelihood-free models can be combined when learning generative models. In the likelihood-free setting, we develop variational inference with synthetic likelihoods that allows us to learn such models.
\item We develop a principled objective function for auto-encoding GANs (\ourgan),\footnote{We use the Greek $\alpha$ prefix for \ourgan, as  AEGAN and most other Latin prefixes seem to have been taken \url{https://deephunt.in/the-gan-zoo-79597dc8c347}.}
 and describe considerations needed to make it work in practice.
\item Evaluation is one of the major challenges in GAN research and we use a battery of evaluation measures to carefully assess the performance of our approach, comparing to DC-GAN, Wasserstein GAN and adversarial-generator-encoders (AGE). We emphasize the continuing challenge of evaluation in implicit generative models and show that our model performs well on these measures.
\end{itemize}
\vspace{-3mm}

\section{Overcoming Intractability in Generative Models}
\label{sect:intractableModels}
\vspace{-2mm}

\mysubsection{Latent Variable Models}
Latent variable models describe a stochastic process by which modeled data is assumed to be generated (and thereby a process by which synthetic data can be simulated from the model distribution). In their simplest form, an unobserved quantity $\vz \sim \zdist$ gives rise to a conditional distribution in the ambient space of the observed data, $\vx \sim \modeldist(\vx | \vz)$. In several recently proposed model families, $\modeldist(\vx |\vz)$ is specified via a generator (or decoder), $\generator(\vz)$, a non-linear function from $\Reals^K\rightarrow\Reals^D$ with parameters $\vtheta$. In this work we consider models with $\vz \sim \N(\bm{0}, \vI)$, unless otherwise specified.

In \textit{implicit latent variable models}, or likelihood-free models, we do not make any further assumptions about the data generating process and set the observation likelihood $\modeldist(\vx|\vz)=\delta(\vx-\generator(\vz))$, which is the model class considered in many simulation-based models, and especially in generative adversarial networks (GANs) \citep{gan}. In \textit{prescribed latent variable models} we make a further assumption of observation noise, and any likelihood function that is appropriate to the data can be used.

In both implicit and prescribed models (such as GANs and VAEs, respectively) an important quantity that describes the quality of the model is the marginal likelihood $\modeldistx$, in which the latent variables $\vz$ have been integrated over. We learn about the parameters $\vtheta$ of the model by minimizing an $f$-divergence between the model likelihood and the true data distribution $\tdistx$, such as the KL-divergence $\KLpq{\tdistx}{\modeldistx}$. But in both types of models, the marginal likelihood is intractable, requiring us to find solutions by which we can overcome this intractability in order to learn the model parameters.

\mysubsection{Generative Adversarial Networks}
One way to overcome the intractability of the marginal likelihood is to never compute it, and instead to learn about the model parameters using a tool that gives us indirect information about it. Generative adversarial networks (GANs) \citep{gan} do this by learning a suitably powerful discriminator that learns to distinguish samples from the true distribution $\tdistx$ and the model $\modeldistx$. The ability of the discriminator (or lack thereof) to distinguish between real and generated data is the learning signal that drives the optimization of the model parameters: when this discriminator is unable to distinguish between real and simulated data, we have learned all we can about the observed data. This is a principle of learning known under various names, including adversarial training \cite{gan}, estimation-by-comparison \citep{gutmann2012noise, gutmann2014statistical}, and unsupervised-as-supervised learning \citep{hastieElements}.

 Let $y=1$ denote a binary label corresponding to data samples from the real data distribution $\vx\sim\tdist$ and $y=0$ for simulated data  $\vx\sim\modeldist$, and a discriminator $\discx=p(y=1|\vx)$ that gives the probability that an input $\vx$ is from the real distribution, with discriminator parameters $\vphi$. At any time point, we update the discriminator by drawing samples from the real data and from the model and minimize the binary cross entropy \eqref{eq:gan_bin_crossentropy}. The generator parameters $\vtheta$ are then updated by maximizing the probability that samples from $\modeldistx$ are classified as real. \citet{gan} suggests an alternative loss in \eqref{eq:gan_alt_loss}, which provides stronger gradients. The optimization is then an alternating minimization w.r.t.~$\vtheta$ and $\vphi$.
 \begin{eqnarray}
&\textrm{\textbf{Discriminator loss:} } \mathbb{E}_{\tdistx}\bigl[- \log \discx\bigr] + \E_{\modeldistx}\bigl[-\log (1 -\discx)\bigr]. \label{eq:gan_bin_crossentropy} \\
& \textrm{\textbf{Generator loss:} } \E_{\modeldistx} [\log (1 - \discx)];
\quad \textrm{\textbf{Alternative loss:} } \E_{\modeldistx}[-\log\discx] \label{eq:gan_alt_loss}
 \end{eqnarray}
GANs are especially interesting as a way of learning in latent variable models, since they do not require inference of the latent variables $\vz$, and are applicable to both implicit and prescribed models. GANs are based on an underlying principle of density ratio estimation  \citep{gan, distinguishability, uehara, implicitgen} and thus provide us with an important tool for overcoming intractable distributions.

\mysubsection{The Density Ratio Trick}
\label{sec:trick}
By introducing the labels $y=1$ for real data and $y=0$ for simulated data in GANs, we re-express the data and model distributions in conditional form, i.e. $\tdistx = p(\vx|y=1)$ for the true distribution, and $\modeldistx = p(\vx|y=0)$ for the model. The \textit{density ratio} $\rphix$ between the true distribution and model distribution can be computed using these conditional distributions as:
\begin{equation}
\rphix=\frac{\tdistx}{\modeldistx}=\frac{p(\vx|y=1)}{p(\vx|y=0)}=\frac{p(y = 1 | \vx)}{p(y = 0 | \vx)}=\frac{\discx}{1-\discx}
\end{equation}

where we used Bayes' rule in the second last step and assumed that the marginal class probabilities are equal, i.e. $p(y=0)=p(y=1)$. This tells us that whenever we wish to compute a density ratio, we can simply draw samples from the two distributions and implement a binary classifier $\discx$ of the two sets of samples. By using the density ratio, GANs account for the intractability of the marginal likelihood by looking only at its relative behavior with respect to the true distribution. This trick only requires samples from the two distributions and never access to their analytical forms, making it particularly well-suited for dealing with implicit distributions or likelihood-free models. Since we are required to build a classifier, we can use all the knowledge we have about building state-of-the-art classifiers.
This trick is widespread \citep{gan, distinguishability, adversarialae, adversarialvb, adversarialmp, huszar2017variational, tran2017deep}. While using class probability estimation is amongst the most popular, the density ratio can also be computed in several other ways including by $f$-divergence minimization and density-ratio matching \citep{sugiyama2012densitybook, implicitgen}.

\mysubsection{Variational Inference}
A second approach for dealing with intractable likelihoods is to approximate them. There are several ways to approximate the marginal likelihood, but one of the most popular is to derive a lower bound to it by transforming the marginal likelihood into an expectation over a new variational distribution $\vardistz$, whose variational parameters $\veta$ can be optimized to ensure that a tight bound can be found. The bound obtained is the popular variational lower bound %
$\mathcal{F}(\vtheta,\veta)$: %
8\begin{align}
 & \log \modeldist(\vx)  = \log \int \modeldist(\vx|\vz)\zdist d\vz  \geq \E_{\vardistz} [ \log \modeldist(\vx|\vz) ] - \KLpq{\vardistz}{\zdist} = \mathcal{F}(\vtheta,\veta).
 \label{eq:free_energy}
 \end{align}
Variational auto-encoders (VAEs) \citep{dlgm, aevb} provide one way of implementing variational inference in which the variational distribution $q$ is represented as an encoder, and the variational and model parameters are  jointly optimized using the pathwise stochastic gradient estimator (also known as the reparameterization trick) \citep{fu2006gradient, aevb, dlgm}. The variational lower bound \eqref{eq:free_energy} is a description applicable to both implicit and prescribed models, and gives us a further tool for dealing with intractable distributions, which is to introduce an encoder to invert the generative process and optimize a lower bound on the marginal likelihood.

\mysubsection{Synthetic Likelihoods}
\label{sect:synthetic_likelihood}
When the likelihood function is unknown, the variational lower bound \eqref{eq:free_energy} cannot directly be used for learning. One further tool with which to overcome this, is to replace the likelihood with a substitute, or \emph{synthetic likelihood} $R(\vtheta)$. The original formulation of the synthetic likelihood \citep{wood2010statistical} is based on a Gaussian assumption, but we use the term here to mean any general substitute for the likelihood that maintains its asymptotic properties. The synthetic likelihood form we use here was proposed by \citet{dutta2016likelihood} for approximate Bayesian computation (ABC).
The idea is to introduce a synthetic likelihood into the likelihood term of \eqref{eq:free_energy} by dividing and multiplying by the true data distribution $\tdistx$:
 \begin{align}
 \E_{\vardistz} \left[ \log \modeldist(\vx|\vz)\right] =
 \E_{\vardistz} \left[ \log \tfrac{\modeldist(\vx|\vz)}{\tdistx}\right]  +  \E_{\vardistz} \left[\log \tdistx\right]
 \label{eq:synthetic_lik}
 \end{align}

The first term in \eqref{eq:synthetic_lik} contains the synthetic likelihood $R(\vtheta) = \tfrac{\modeldist(\vx|\vz)}{\tdistx}$. Any estimate of the ratio $R(\vtheta)$ is an estimate of the likelihood since they are proportional (and the normalizing constant is independent of $\vtheta$). %
Wherever an intractable likelihood appears, we can instead use this ratio.
The synthetic likelihood can be estimated using the density ratio trick by training a discriminator to distinguish between samples from the marginal $\tdistx$ and the conditional  $\modeldist(\vx | \vz)$ where $
\vz$ is drawn from $\vardistz$.
The second term in \eqref{eq:synthetic_lik} is independent of $\vtheta$ and can be ignored for optimization purposes.

\section{A Fusion of Variational and Adversarial Learning}
\vspace{-2mm}
GANs and VAEs have given us useful tools for learning and inference in generative models and we now use these tools to build new hybrid inference methods. The VAE forms our generic starting point, and we will gradually transform it to be more GAN-like.

\mysubsection{Implicit Variational Distributions} The major task in variational inference is the choice of the variational distribution $\vardistz$. Common approaches, such as mean-field variational inference, assume simple distributions like a Gaussian, but we would like not to make a restrictive choice of distribution. If we treat this distribution as implicit---we do not know its distribution but are able to generate from it---then we can use the density ratio trick to  replace the KL-divergence term in \eqref{eq:free_energy}.
\begin{align}
-\textrm{KL}[\vardistz \| p(\vz)] = \E_{\vardistz}  \left[ \log \frac{p(\vz)}{\vardistz} \right] \approx \E_{\vardistz}  \left[ \log \frac{\codediscz}{1-\codediscz} \right]. \label{eq:KLratio}
\end{align}
We will thus introduce a latent classifier $\codediscz$ that discriminates between latent variables $\vz$ produced by an encoder network and variables sampled from a standard Gaussian distribution. For optimization, the expectation in \eqref{eq:KLratio} is evaluated by Monte Carlo integration. Replacing the KL-divergence with a discriminator was first proposed by \citet{adversarialae}, and a similar idea was used by \citet{adversarialvb} for adversarial variational Bayes.

\mysubsection{Likelihood Choice}
If we make the explicit choice of a likelihood $p_{\vtheta}(\vx | \vz)$ in the model, the we can substitute our chosen likelihood into \eqref{eq:free_energy}. We choose a zero-mean Laplace distribution $\modeldist(\vx|\vz)\propto\exp(-\lambda||\vx-\generator(\vz)||_1)$ with scale parameter $\lambda$, which corresponds to using a variational auto-encoder with an $L_1$ reconstruction loss; this is a highly popular choice and used in many related auto-encoder GAN variants, such as AGE, BEGAN, cycle GAN and PPGN \citep{age, began, ppgn, cyclegan}.

In GANs the effective likelihood is unknown and intractable. We can again use our tools for intractable inference by replacing the intractable likelihood by its synthetic substitute. Using the synthetic likelihood  \eqref{eq:synthetic_lik} introduces a new synthetic-likelihood classifier $\discx$ that discriminates between data sampled from the conditional and marginal distributions of the model.
The reconstruction term $\E_{\vardistz} [ \log \modeldist(\vx|\vz)] $ in \eqref{eq:free_energy} can be either:
\begin{align}
\E_{\vardistz} \left[-\lambda||\vx-\generator(\vz)||_1\right]  \qquad \textrm{or} \qquad  \E_{\vardistz} \left[\log\frac{\disc(\generator(\vz))}{1-\disc(\generator(\vz))}\right].
\label{eq:fe_loglik_synthetic}
\end{align}

These two choices have different behaviors. Using the synthetic discriminator-based likelihood means that this model will have the ability to use the adversarial game to learn the data distribution, although it may still be subject to mode-collapse. This is where an explicit choice of likelihood can be used to ensure that we assign mass to all parts of the output support and prevent collapse. When forming a final loss we can make use of a weighted sum of the two to get the benefits of both types of behavior.

\mysubsection{Hybrid Loss Functions} An hybrid objective function that combines all these choices is:
\begin{align}
\mathcal{L}(\vtheta,\veta) & = \E_{\vardistz}  \left[ -\lambda||\vx-\generator(\vz)||_1 + \log \frac{\disc(\generator(\vz))}{1-\disc(\generator(\vz))} + \log \frac{\codediscz}{1-\codediscz} \right]   \label{eq:final_loss}
\end{align}
We are required to build four networks: the classifier $\discx$ is  trained to discriminate between reconstructions from an auto-encoder and real data points;  a second classifier is trained to discriminate between latent samples produced by the encoder and samples from a standard Gaussian; we must implement the deep generative model $\generator(\vz)$, and also the encoder network $\vardistz$, which can be implemented using any type of deep network.
The density-ratio estimators $\disc$ and $\codedisc$ can be trained using any loss for density ratio estimation described in section~\ref{sec:trick}, hence  their loss functions are not shown in  \eqref{eq:final_loss}.  We refer to training using  \eqref{eq:final_loss}  as \ourgan. Our algorithm alternates between updates of the parameters of the generator $\vtheta$,  encoder $\veta$, synthetic likelihood discriminator $\vphi$,  and the latent code discriminator $\vomega$; see algorithm \ref{alg:pseudocode}.

\mysubsection{Improved Techniques}
\label{sect:impr_techniques}
Equation \eqref{eq:final_loss} provides a principled starting point for optimization based on losses obtained by the combination of insights from VAEs and GANs. To improve the stability of optimization and speed of learning we make two modifications. Firstly, following the insights from \citet{implicitgen}, we consider the reverse KL loss formulation for both the latent discriminator and the synthetic likelihood discriminator, where we replace $-\log(1-\disc)$ with $\log\disc -\log(1-\disc)$ while training the generator as it provides non-saturating gradients. The minimization of the generator parameters becomes:
\begin{equation}
\textrm{\textbf{Generator Loss:} }  \E_{\vardistz} \Bigl[ \lambda||\vx-\generator(\vz)||_1 - \log{\disc(\generator(\vz))} + \log{(1 - \disc(\generator(\vz)))} \Bigr],
\end{equation}
which shows that we have are using the GAN updates for the generator, with the addition of a reconstruction term, that discourages mode collapse as $\generator$ needs to be able to reconstruct every input $\vx$.

Secondly, we found that passing the samples to the discriminator as fake samples, in addition to the reconstructions, helps improve performance. One way to justify the use of samples is to apply Jensen's inequality, that is,
$ \log \modeldist(\vx)  = \log \int \modeldist(\vx|\vz)\zdist d\vz  \geq \E_{\zdist} [ \log \modeldist(\vx|\vz) ]$, and replace this with a synthetic likelihood, as done for reconstructions. Instead of training two separate discriminators, we train a single discriminator which treats samples and reconstructions as fake, and $\tdist$ as real. %

\vspace{-2mm}
\section{Related work}\label{sec:related}
\begin{figure}[t]
\centering
\begin{subfigure}[DCGAN]{
\includegraphics[height=3.5cm]{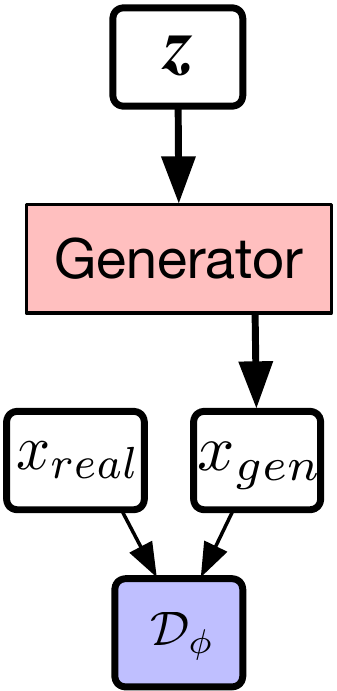}
}\end{subfigure}
\hspace{5mm}
\begin{subfigure}[AGE]{
\includegraphics[height=3.5cm]{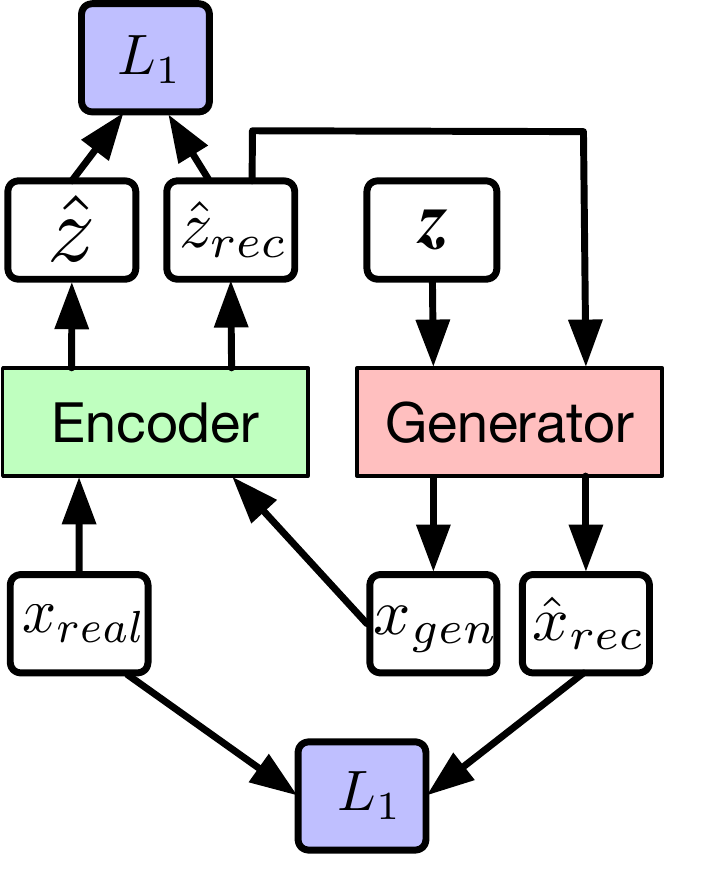}
}\end{subfigure}
\hspace{5mm}
\begin{subfigure}[\ourgan]{
\includegraphics[height=3.5cm]{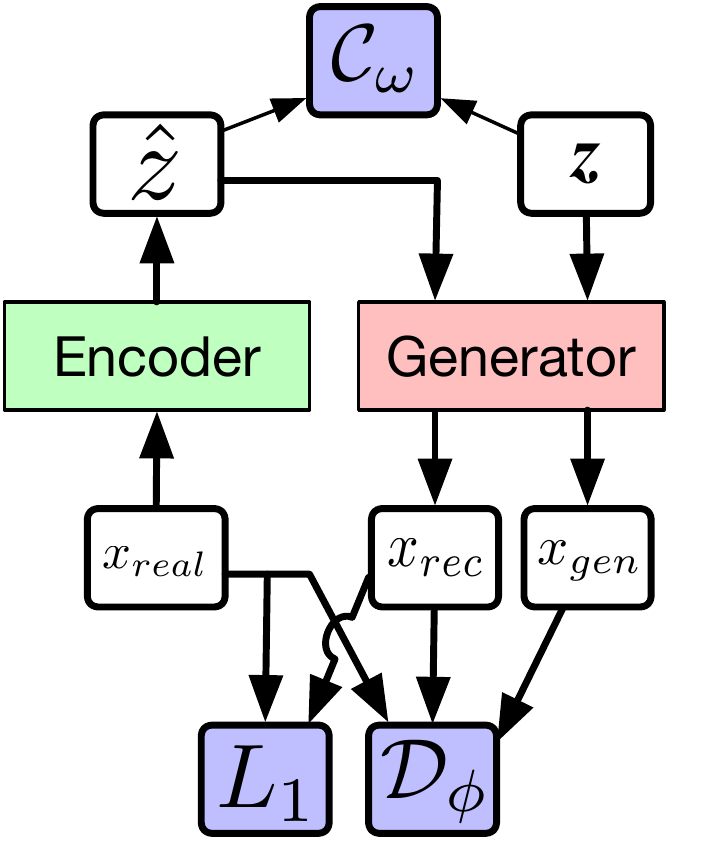}
}\end{subfigure}
\caption{{Architectures for the three models used for comparison. (WGAN is similar to DCGAN.)}}
\label{fig:model}
\end{figure}

\vspace{-2mm}
Figure~\ref{fig:model} summarizes our architecture and the architectures we compare with in the experimental section.
Hybrids of VAEs and GANs can be classified by whether the density ratio trick is applied only to likelihood, prior approximation or both.
Table~\ref{tab:models} reveals the connections to related approaches (see also \citep[Table 1]{huszar2017variational}).
DCGAN \citep{dcgan} and WGAN-GP \citep{improvedwgan} are pure GAN variants; they do not use an auto-encoder loss nor do they do inference. WGAN-GP shares the attributes of DCGAN, except that it uses a critic that approximates the Wasserstein distance \citep{wgan} instead of a density ratio estimator. AGE uses an approximation of KL term, however it does not use a synthetic likelihood, but instead uses observed likelihoods - reconstruction losses - for both latent codes and data. The adversarial component of AGE arises form the opposing goals of the encoder and decoder: the encoder tries to compress data into codes drawn from the prior, while compressing samples into codes which do not match the prior; at the same time the decoder wants to generate samples that when encoded by the encoder will generate codes which match the prior distribution. VAE uses the observation likelihood and an analytic KL term, however it tends to produce blurry images, hence we do not consider it here. To solve the blurriness issue, VAE-GAN change the VAE loss function by replacing the observed likelihood on pixels with an adversarial loss together with a reconstruction metric in discriminator feature space. Unlike our work, VAE-GAN still uses the analytical KL loss to minimize the distance between the prior and the posterior of the latents, and they do not discuss the connection to density ratio estimation. Similar to VAE-GAN, \citet{deepsim} replace the observed likelihood term in the variational lower bound with a weighted sum of a feature matching loss (here the features matched are those of a pre-trained classifier) and an adversarial loss, but instead of using the analytical KL, they use a numerical approximation. We explore the same approximation (also used by AGE) in Section \ref{app:code_empirical_kl} in the Appendix. By not using a pre-trained classifier or a feature matching loss, \ourgan~is trained end-to-end, completely unsupervised and maximizes a lower bound on the true data likelihood.

ALI \citep{ali}, BiGAN \citep{bigan} perform inference by creating an adversarial game between the encoder and decoder via a discriminator that operates on $\vx,\vz$ space. The discriminator learns to distinguish between input-output pairs of the encoder (where $\vx$ is a sample from the data distribution and $\vz$ is a sample from the conditional posterior $\vardistz$) and decoder (where $\vz$ is a sample from the latent prior and $\vx$ is a sample from the conditional $\modeldist(\vx|\vz)$). Unlike \ourgan~, their approach operates jointly, without exploiting the structure of the model.
Cycle-GAN \citep{cyclegan} was proposed for image-to-image translation, but applying the underlying \emph{cycle consistency} principle to {image-to-code} translation reveals an interesting connection with \ourgan. %
This method has become popular for image-to-image translation, with similar approaches having proposed\citep{discogan}.
Recall that in $\vx$ space, we both use a pointwise reconstruction term $\Vert\vx-\vxhat\Vert_1$ term as well as a loss to match the distributions of $\vx$ and $\vxhat$. In $\vz$ space, we only match the distributions of $\vz$ and $\vzhat$ in \ourgan. Adding  pointwise code reconstruction loss $||\vz-\vzhat||$  would make it similar to CycleGAN. We note however that the CycleGAN authors used the least square GAN loss, while the traditional GAN loss needs to be used to obtain the variational lower bound in \eqref{eq:free_energy}.

In mode regularized GANs (MRGANs) \citep{moderegularizedgan} the generator is part of an auto-encoder, hence it learns how to produce reconstructions from the posterior over latent codes and also independently learns how to produce samples from codes drawn from the prior over latents. MRGANs employ two discriminators, one to distinguish between data and reconstructions and one to distinguish between data and samples. As described in Section \ref{sect:impr_techniques}, in \ourgan~we also pass both samples and reconstructions through the discriminator (which learns to distinguish between them and data). However, we only need one discriminator, as we explicitly match the latent prior and the latent posterior given by the model using KL term in \eqref{eq:free_energy}, which encourages the distributions of reconstructions and sample to be similar.

\begin{table}[ht]%
\begin{center}
\resizebox{0.95\columnwidth}{!}{
\begin{tabular}{l | cc | ccc}
\toprule
 Algorithm & \multicolumn{2}{|c|}{Likelihood} & \multicolumn{3}{c}{Prior}  \\
& Observer  &  Ratio estimator ("synthetic") & KL (analytic) & KL (approximate) &  Ratio estimator \\
 \midrule
 VAE & \checkmark & & \checkmark & & \\
DCGAN &  & \checkmark & & & \\
  VAE-GAN & \checkmark & * & \checkmark & & \\
AGE & \checkmark & &  & \checkmark & \\
\ourgan~(ours) & \checkmark & \checkmark &  &  & \checkmark \\
 \bottomrule
\end{tabular}
}
\end{center}
\caption{Comparison of different approaches for training generative latent variable models.
}
\label{tab:models}
\end{table}

\vspace{-2mm}
\section{Evaluation metrics}\label{sec:metrics}
\vspace{-2mm}
Evaluating generative models is challenging \citep{noteevaluation}. In particular, evaluating GANs is difficult due to the lack of likelihood. %
Multiple proxy metrics have been proposed, and we explore some of them in this work and assess their strengths and weaknesses in the experiments section.

\evalsubsection{Inception score}
 The inception score was proposed by \citet{improvedgan} and has been widely adopted since. The inception score uses a pre-trained neural network classifier to  capture to two desirable properties of generated samples: highly classifiable and diverse with respect to class labels. It does so by computing the average of the KL divergences between the conditional label distributions of samples (expected to have low entropy for easily classifiable samples) and the marginal distribution obtained from all the samples (expected to have high entropy if all classes are equally represented in the set of samples). As the name suggests, the classifier network used to compute the inception score was originally an Inception network \citep{inception} trained on the ImageNet dataset. For comparison to previous work, we  report scores using this network.
However, when reporting CIFAR-10 results we  also report metrics obtained using a VGG style convolutional neural network, trained on the same dataset, which obtained 5.5\% error (see section~\ref{app:cifar_10_net} in the details on this network).

\evalsubsection{Multi-scale structural similarity (MS-SSIM)}
 The inception score fails to capture mode collapse inside a class: the inception score of a model that generates the same image for a class and the inception score of a model that is able to capture diversity inside a class are the same. To address this issue, \citet{acgan} assess the similarity between class-conditional generated samples using MS-SSIM \citep{wang2003multiscale}, an image similarity metric that has been shown to correlate well with human judgement.
 MS-SSIM ranges between 0.0 (low similarity) and 1.0 (high similarity). By computing the average pairwise MS-SSIM score between images in a given set, we can determine how similar the images are, and in particular, we can compare with the similarity obtained on a reference set (the training set, for example).  Since our models are not class conditional, we only used MS-SSIM to evaluate models on CelebA \citep{celeba}, a dataset of faces, since the variability of the data there is smaller. For datasets with very distinct labels, using MS-SSIM would not give us a good metric, since there will be high variability between classes. We report \emph{sample diversity score} as 1-MSSSIM. The reported results on this metric need to be seen relative to the diversity obtained on the input dataset: too much diversity can mean failure to capture the data distribution. To illustrate this, we computed the diversity on images from the input dataset to which we add normal noise, and it is higher than the diversity of the original data. We report this value as another baseline for this metric.

\evalsubsection{Independent Wasserstein critic}
\citet{ivogan} proposed training an independent Wasserstein GAN critic to distinguish between held out validation data and generated samples.\footnote{\citet{ivogan} used the original WGAN \citep{wgan}, whereas we use improved WGAN-GP proposed in \citep{improvedwgan}.}
This metric measures both overfitting and mode collapse: if the generator memorizes the training set, the critic trained on validation data will be able to distinguish between samples and data; if mode collapse occurs, the critic will have an easy task distinguishing between data and samples. The Wasserstein distance does not saturate when the two distributions do not overlap \citep{wgan}, and the magnitude of the distance represents how easy it is for the critic to distinguish between data and samples.
To be consistent with the other metrics, we report the negative of the Wasserstein distance between the test set and generator, hence higher values are better.
Since the critic is trained independently for evaluation only, and thus does not affect the training of the generator, this evaluation technique can be used irrespective of the training criteria used \citep{ivogan}.
To ensure that the independent critic does not overfit to the validation data, we only start training it half way through the training of our model and examined the learning curves during training (see Appendix~\ref{app:improved_critic} in the supplementary material for learning curves).

\section{Experiments}\label{sec:experiments}
\vspace{-2mm}
To better understand the importance of autoencoder based methods in the GAN landscape, we implemented and compared the proposed \ourgan~with %
another hybrid model, AGE,
as well as pure GAN variants such as DCGAN and WGAN-GP, across three datasets: \colormnist~\citep{unrolledgan}, CelebA \citep{celeba} and CIFAR-10 \citep{cifar10}. We complement the visual inspection of samples with a battery of numerical test using the metrics above to get an insight of both on the models and on the metrics themselves. For a comprehensive analysis, we report both the best values obtained by each algorithm, as well as the quartiles obtained by each hyperparameter sweep for each model, to assess the sensitivity to hyperparameters. On all metrics, we report box plot for all the hyperparameters we considered with the best 10 jobs indicated by black circles  (for Inception Scores and Independent Wasserstein critic, higher is better; for sample diversity, the best reported jobs are those with the smallest distance from the reference value computed on the test set).
To the best of our knowledge, we are the first to do such an analysis of the GAN landscape.

For details of the training procedure used in all our experiments, including the hyperparameter sweeps, we refer to Appendix~\ref{app:hyperparameters} in the supplementary material.
Note that the models considered here are all unconditional and do not make use of label information, hence it is not appropriate to compare our results with those obtained using conditional GANs \citep{acgan} and semi-supervised GANs \citep{improvedgan}.

\begin{figure*}[ht]
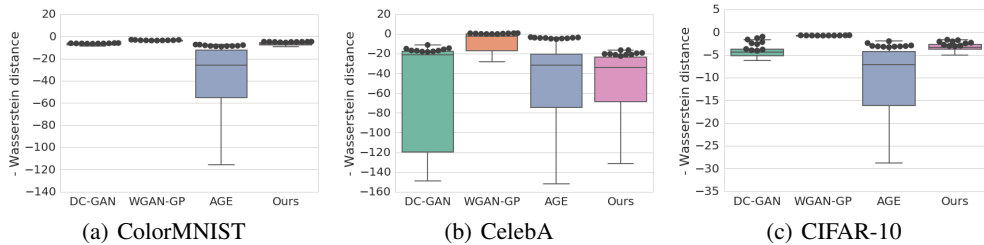

\begin{center}
\begin{subfigure}[\colormnist]{
\includegraphics[width=0.3\textwidth]{colour-mnist-wgan-critic}
\label{fig:wcritics-colormnist}
}\end{subfigure}
\begin{subfigure}[CelebA]{
\includegraphics[width=0.3\textwidth]{celeba-64-wgan-critic}
\label{fig:wcritics-celeba}
}\end{subfigure}
\begin{subfigure}[CIFAR-10]{
\includegraphics[width=0.3\textwidth]{cifar-wgan-critic}
\label{fig:wcritics-cifar}
}\end{subfigure}
\caption{{Negative Wasserstein distance estimated using an independent Wasserstein critic. The metric captures overfitting to the training data and low quality samples. Higher is better.}
}
\label{fig:wcritics-all}
\end{center}
\vspace{-4mm}
\end{figure*}

\begin{figure*}[ht]
\begin{center}
\includegraphics[width=0.24\textwidth]{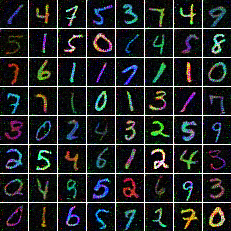} \addhspacesmall
\includegraphics[width=0.24\textwidth]{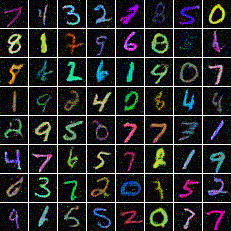} \addhspacesmall
\includegraphics[width=0.24\textwidth]{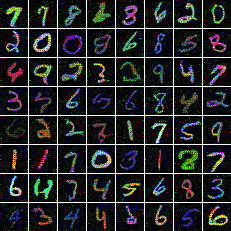} \addhspacesmall
\includegraphics[width=0.24\textwidth]{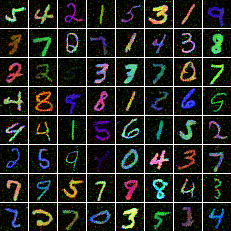} \addhspacesmall
\caption{
Best samples on \colormnist~(L-to-R): samples from DCGAN, WGAN-GP, AGE and the proposed variant \ourgan, according to visual inspection.
}
\label{fig:colormnist:samples}
\end{center}
\vspace{-3mm}
\end{figure*}

\evalsubsection{Results on \colormnist~}
We compare the values of an independent Wasserstein critic in Figure \ref{fig:wcritics-colormnist}, where higher values are better. %
On this metric, most of hyperparameters tried achieve a higher value than the best DC-GAN results. This is supported by the generated samples shown in Figure \ref{fig:colormnist:samples}. However, WGAN-GP produces the samples rated best by this metric.

\begin{figure*}[t]
\begin{center}
\includegraphics[width=0.24\textwidth]{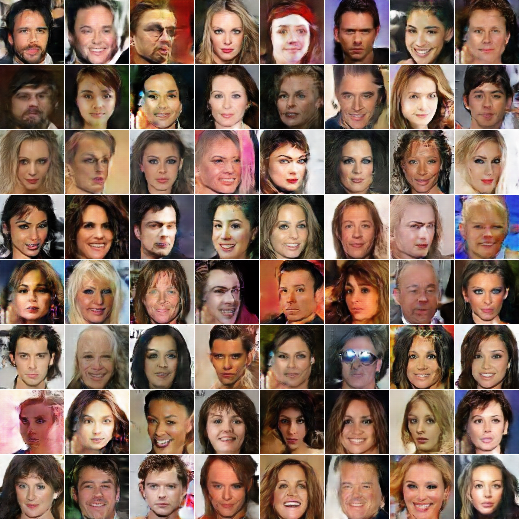} \addhspacesmall
\includegraphics[width=0.24\textwidth]{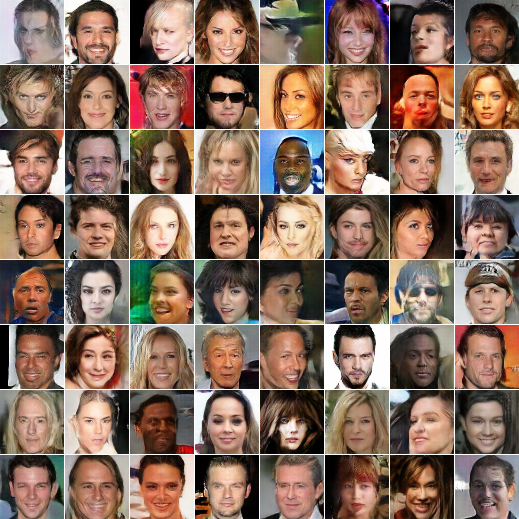} \addhspacesmall
\includegraphics[width=0.24\textwidth]{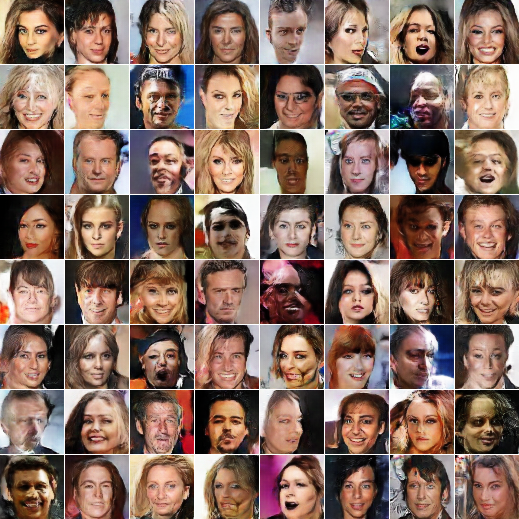} \addhspacesmall
\includegraphics[width=0.24\textwidth]{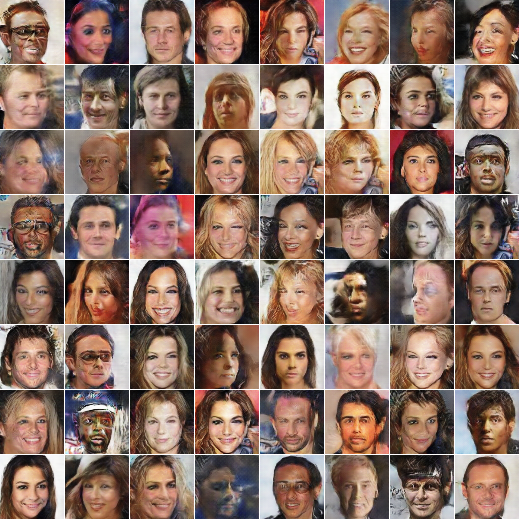} \addhspacesmall\caption{
Best samples on CelebA (L-to-R): samples from DCGAN, WGAN-GP, AGE and \ourgan, according to visual inspection.
See Figure \ref{fig:celeba:samples_big} in Appendix~\ref{app:model:samples} for a higher resolution version.
}
\label{fig:celeba:samples}
\end{center}
\vspace{-2mm}
\end{figure*}

\begin{figure*}[t]
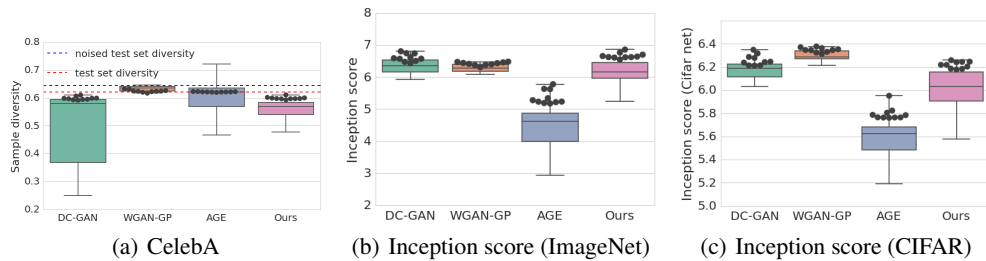

\begin{center}
\begin{subfigure}[CelebA]{
\includegraphics[width=0.3\textwidth]{celeba-64-mssim}
\label{fig:mssim_inception:celeba}
}\end{subfigure}
\begin{subfigure}[Inception score (ImageNet)]{
\includegraphics[width=0.3\textwidth]{cifar-inception-score-inception}
\label{fig:mssim_inception:inceptionnet}
}\end{subfigure}
\begin{subfigure}[Inception score (CIFAR)]{
\includegraphics[width=0.3\textwidth]{cifar-inception-score-cifar}
\label{fig:mssim_inception:cifarnet}
}\end{subfigure}
\caption{
Left plot shows sample diversity results on CelebA (performance should be compared to the test set baseline).  Middle plot: Inception score results on CIFAR-10. Right most plot shows Inception score computed using a VGG style network trained on CIFAR-10. For Inception scores, high values are better. As a reference benchmark, we also compute these scores using samples from test data split; diversity: 0.621, inception score: 11.25, inception score (VGG net trained on CIFAR-10): 9.18.
}
\label{fig:mssim_inception}
\end{center}
\vspace{-2mm}
\end{figure*}

\evalsubsection{Results on CelebA}
The CelebA dataset consists of $64\times64$ pixel images of faces of celebrities. We show samples from the four models in Figure \ref{fig:celeba:samples}. We also compare the models using the independent Wasserstein critic in Figure \ref{fig:wcritics-celeba} and sample diversity score in Figure \ref{fig:mssim_inception:celeba}. \ourgan~is competitive with WGAN-GP and AGE, but has a wider spread than the WGAN-GP model, which produces the best results.

Unlike WGAN and DCGAN, an advantage of \ourgan~and AGE is the ability to reconstruct inputs. Appendix~\ref{app:reconstructions} shows that \ourgan~produces better reconstructions than AGE.

\begin{figure*}[ht]
\begin{center}
\includegraphics[width=0.24\textwidth]{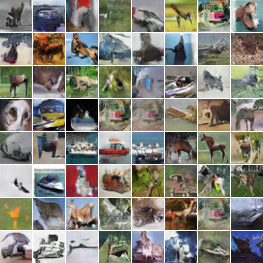} \addhspacesmall
\includegraphics[width=0.24\textwidth]{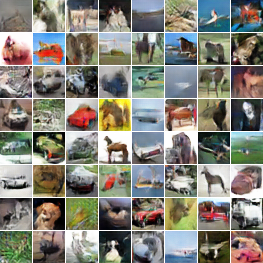} \addhspacesmall
\includegraphics[width=0.24\textwidth]{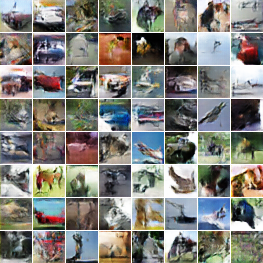} \addhspacesmall
\includegraphics[width=0.24\textwidth]{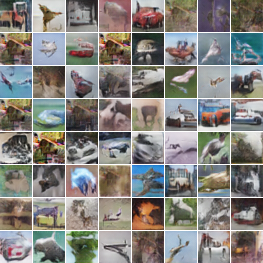} \addhspacesmall
\caption{Best samples on CIFAR-10 (L-to-R): DC-GAN, WGAN-GP, AGE and \ourgan, according to visual inspection. For AGE and \ourgan~reconstructions on CIFAR-10, see Appendix~\ref{app:reconstructions}.
}
\label{fig:cifar:samples}
\end{center}
\end{figure*}

\evalsubsection{Results on CIFAR-10}
We show samples from the various models in Figure \ref{fig:cifar:samples}. We evaluate \ourgan~using the independent critic, shown in Figure \ref{fig:wcritics-cifar}, where WGAN-GP is the best performing model. We also compare the ImageNet-based inception score  in Figures \ref{fig:mssim_inception:inceptionnet}, where it has the best performance, and with the CIFAR-10 based inception score in Figure \ref{fig:mssim_inception:cifarnet}. While our model produces the best Inception score result on the ImageNet-based inception score, it has wide spread on the CIFAR-10 Inception score, where WGAN-GP both performs best, and has less hyperparameter spread.
This shows that the two metrics widely differ, and that evaluating CIFAR-10 samples using the ImageNet based inception score can lead to erroneous conclusions.
To understand more of the importance of the model used to evaluate the Inception score, we looked at the relationship between the Inception score measured with the Inception net trained on ImageNet (introduced by \citep{improvedgan}) and the VGG style net trained on CIFAR-10, the same dataset on which we train the generative models. We observed that 15\% of the jobs in a hyperparameter sweep were ranked as being in the top 50\% by the ImageNet Inception score while ranked in the bottom 50\% by the CIFAR-10 Inception score. Hence, using the Inception score of a model trained on a different dataset than the generative model is evaluated on can be misleading when ranking models.

The best reported ImageNet-based inception score on CIFAR for unsupervised models is $7.72\pm0.13$ by DFM-GAN \citep{davidgan}, who also report
$5.34\pm0.05$ for ALI \citep{ali}, however these are trained on different architectures and may not be directly comparable.

\evalsubsection{Experimental insights}
Irrespective of the algorithm used, we found that two factors can contribute significantly to the quality of the results:
\vspace{-1mm}
\begin{itemize}[leftmargin=2ex,topsep=0pt,itemsep=-1ex,partopsep=1ex,parsep=1ex]
\item \emph{The network architectures}. We noticed that the most decisive factor in the lies in the architectures chosen for the discriminator and generator. We found that given enough capacity, DCGAN (which uses the traditional GAN \citep{gan})  %
can be very robust, and does not suffer from obvious mode collapse on the datasets we tried. All models reported are sensitive to changes in the architectures, with minor changes resulting in catastrophic mode collapse, regardless of other hyperparameters.
\item \emph{The number of updates performed by the individual components of the model}. For DCGAN, we update the generator twice for each discriminator update following \url{https://github.com/carpedm20/DCGAN-tensorflow}; we found it stabilizes training and produces significantly better samples, contrary to GAN theory which suggests training discriminator multiple times instead.
 Our findings are also consistent with the updates performed by the AGE model, where the generator is updated multiple times for each encoder update. Similarly, for \ourgan, we update the encoder (which can be seen as the latent code generator) and the generator twice for each discriminator and code discriminator update. On the other hand, for WGAN-GP, we update the discriminator 5 times for each generator update following \cite{wgan,improvedwgan}.
\end{itemize}

While the independent Wasserstein critic does not directly measure sample diversity, we notice a high correlation between its estimate of the negative Wasserstein distance and sample similarity (see Appendix~\ref{app:wgan_mssim}). Note however that the measures are not perfectly correlated, and if used to rank the best performing jobs in a hyperparameter sweep they give different results.

\section{Discussion}

In this paper we have combined the variational lower bound on the data likelihood with the density ratio trick, allowing us to better understand the connection between variational auto-encoders and generative adversarial networks. From the newly introduced lower bound on the likelihood we derived a new training criteria for generative models, named \ourgan. \ourgan~combines an adversarial loss with a data reconstruction loss. This can be seen in two ways: from the VAE perspective, it can solve the blurriness of samples via the (learned) adversarial loss; from the GAN perspective, it can solve mode collapse by grounding the generator using a perceptual similarity metric on the data - the reconstruction loss. In a quest to understand how \ourgan~compares to other GAN models (including auto-encoder based ones), we deployed a set of metrics on 3 datasets as well as compared samples visually. While the picture of evaluating GANs is far from being completed, we show that the metrics employed are complementary and assess different failure modes of GANs (mode collapse, overfitting to the training data and poor learning of the data distribution).

The prospect of marrying the two approaches (VAEs and GANs) comes with multiple benefits: auto-encoder based methods can be used to reconstruct data and thus can be used for inpainting \citep{ppgn} \citep{inpainting}; having an inference network allows our model to be used for representation learning \citep{bengio2013representation}, where we can learn disentangled representations by choosing an appropriate latent prior. We thus believe VAE-GAN hybrids such as \ourgan~can be used in unsupervised, supervised and reinforcement learning settings, which leads the way to directions of research for future work.

\clearpage
\newpage

\textbf{Acknowledgements.} We thank Ivo Danihelka and Chris Burgess for helpful feedback and discussions.
\vspace{-5mm}
\bibliography{paper}%
\bibliographystyle{abbrvnat}

\clearpage

\appendix
{\centerline{\Large{\textbf{Supplementary material}}}}
\section{Model Samples}\label{app:model:samples}
\begin{figure*}[ht]
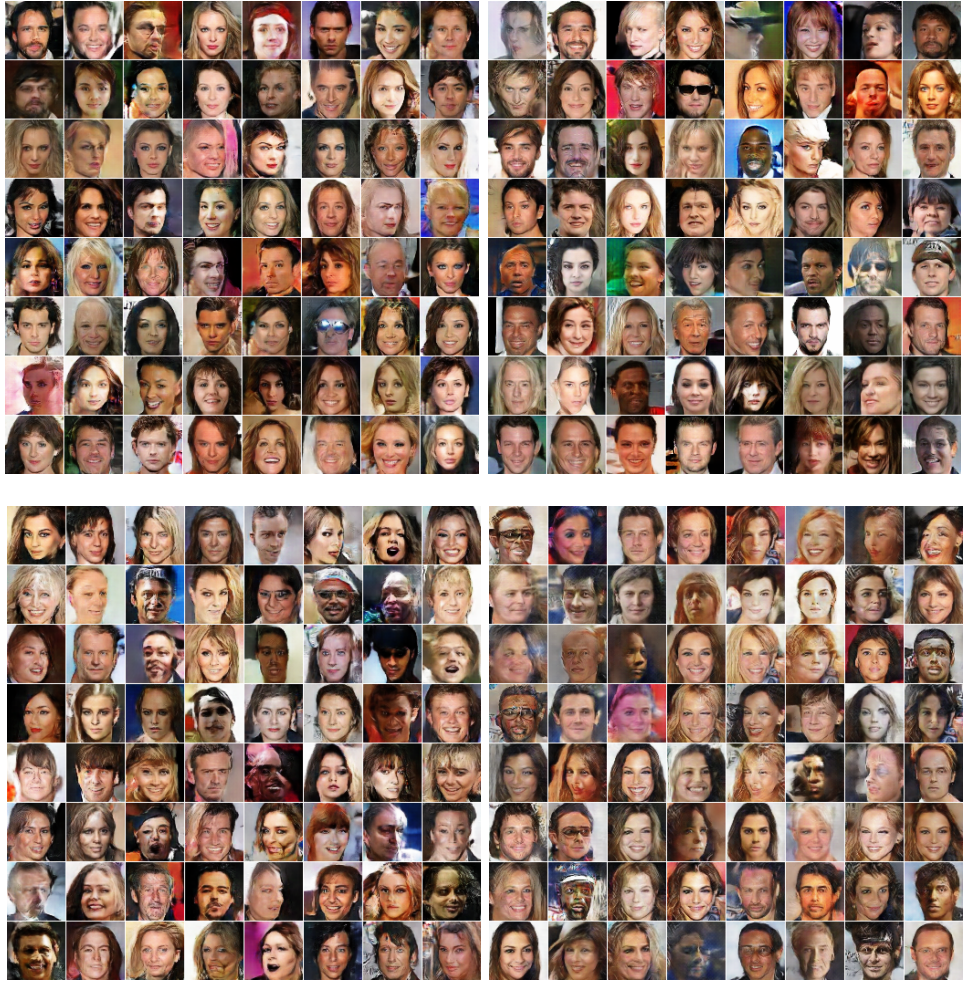

\begin{center}
\includegraphics[width=0.45\textwidth]{celeba_dcgan_latest} \addhspacesmall
\includegraphics[width=0.45\textwidth]{celeba_wgan_latest} \addhspacesmall
\vspace{4mm}\hspace{5mm}\\
\includegraphics[width=0.45\textwidth]{celeba_age_latest} \addhspacesmall
\includegraphics[width=0.45\textwidth]{celeba_our_latest} \addhspacesmall\caption{
Best samples on CelebA according to visual inspection shown in Figure \ref{fig:celeba:samples}. \textit{Top row}: (left) DCGAN (right) WGAN-GP. \textit{Bottom row}: (left) AGE (right) \ourgan.
}
\label{fig:celeba:samples_big}
\end{center}
\end{figure*}

Figure~\ref{fig:celeba:samples_big} shows larger-sized versions of the samples in Figure~\ref{fig:celeba:samples} in the main text.

\section{Pseudocode}\label{app:pseudocode}
The overall training procedure is summarized in Algorithm~\ref{alg:pseudocode}.
\begin{algorithm*}
\caption{Pseudocode for \ourgan}
\label{alg:pseudocode}
\begin{algorithmic}[1]
\State Initialize parameters of generator $\vtheta$, encoder (variational distribution) $\veta$, discriminator $\vphi$ and code discriminator $\vomega$ randomly.
\State Let $\vzhat\sim\vardistz$ denote a sample from the encoding variational distribution $\vardistz$ and $\vxhat = \generator(\vzhat)$ denote the `reconstruction' of $\vx$ using $\vzhat$.
\State Let $R_{\disc}(\vx) = - \log{\disc(\vx)} + \log{(1- \disc(\vx))}$
\State Let $R_{\codedisc}(\vz) = - \log{\codedisc(\vz)} + \log{(1- \codedisc(\vz))}$
\For{$\mathsf{iter}=1:\mathsf{max\_iter}$}
\State Update encoder (variational distribution) $\veta$ by minimizing
\newline\indent\algcomment{reconstruction and generation loss from the code discriminator}
\begin{align}
& \E_{\tdistx}  \E_{\vardistz} \Big[ \lambda||\vx-\generator(\vz)||_1 \bigr]  + R_{\codedisc}({\vz})  \Bigr] \\
& \approx \E_{\tdistx} \Big[ \lambda||\vx-\vxhat||_1 \bigr] + R_{\codedisc}({\vzhat})\Bigr]
\end{align}

\State Update generator $\vtheta$ by minimizing  \algcomment{reconstruction and generation loss}
\begin{align}
& \mathbb{E}_{\tdistx}\E_{\vardistz} \bigl[ \lambda||\vx-\generator(\vz)||_1 + R_{kl}(\vz)\bigr] + \E_{\zdist} \bigl[ R_{\disc}(\generator(\vz))\bigr] \\
& \approx \mathbb{E}_{\tdistx}  \bigl[ \lambda||\vx-\vxhat||_1 + R_{\disc}(\vxhat) \bigr] + \E_{\zdist} \bigl[ R_{\disc}(\generator(\vz)) \bigr]
\end{align}

\State Update discriminator $\vphi$ by minimizing
\newline\indent \algcomment{treat $\tdistx$ as real, reconstructions and generated samples as fake}
\begin{align}
& \mathbb{E}_{\tdistx}\bigl[-2 \log\discx  - \E_{\vardistz}\log\bigl(1-\disc(\generator(\vz))\bigr)
\bigr] + \E_{\zdist} \bigl[- \log\bigl(1-\disc(\generator(\vz))\bigr)\bigr] \\
&\approx \mathbb{E}_{\tdistx}\bigl[-\log\discx  - \log\bigl(1-\disc(\vxhat)\bigr)
\bigr] + \E_{\zdist} \bigl[- \log\bigl(1-\disc(\generator(\vz))\bigr)\bigr]
\end{align}

\State Update code discriminator $\vomega$ by minimizing
\newline\indent\algcomment{treat $p(\vz)$ as real and codes from variational distribution as fake}
\begin{align}
& \E_{\tdistx} \E_{\vardistz} \bigl[ -\log(1-\codediscz)\bigr] + \E_{\zdist} \bigl[-\log\codediscz\bigr] \\
& \approx  \E_{\tdistx}  \bigl[-\log(1-\codedisc(\vzhat))\bigr] + \E_{\zdist} \bigl[ -\log(\codediscz)\bigr]
\end{align}
\EndFor
\end{algorithmic}
\end{algorithm*}

\section{Reconstructions \label{app:reconstructions}}
\begin{figure*}[htbp]
\begin{center}
\begin{subfigure}[AGE]{
\includegraphics[width=0.45\textwidth]{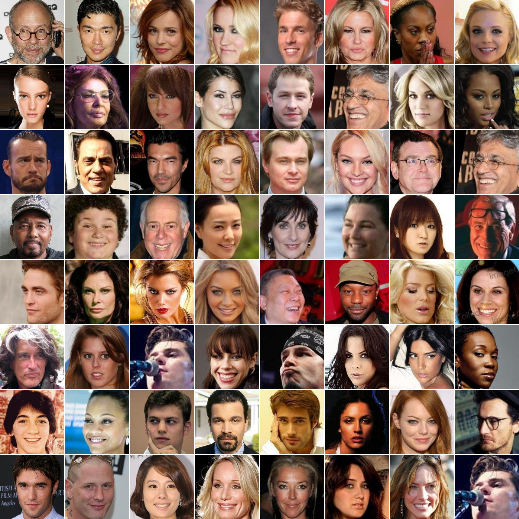} \addhspacesmall
\includegraphics[width=0.45\textwidth]{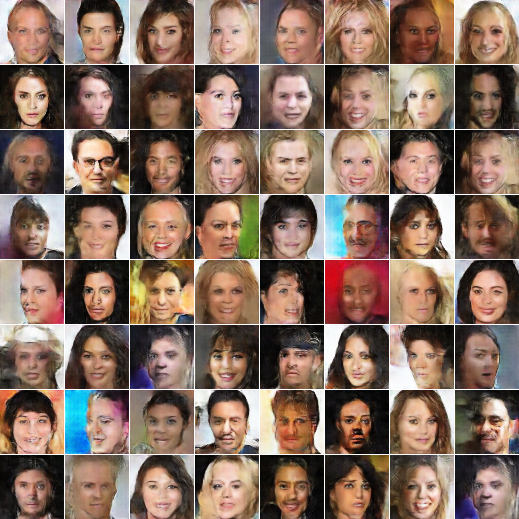} \addhspacesmall
\label{fig:celeba:reconstructions:age}
}\end{subfigure}
\begin{subfigure}[\ourgan]{
\includegraphics[width=0.45\textwidth]{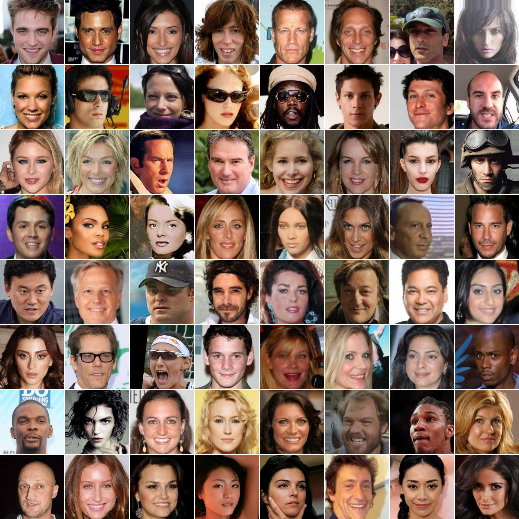} \addhspacesmall
\includegraphics[width=0.45\textwidth]{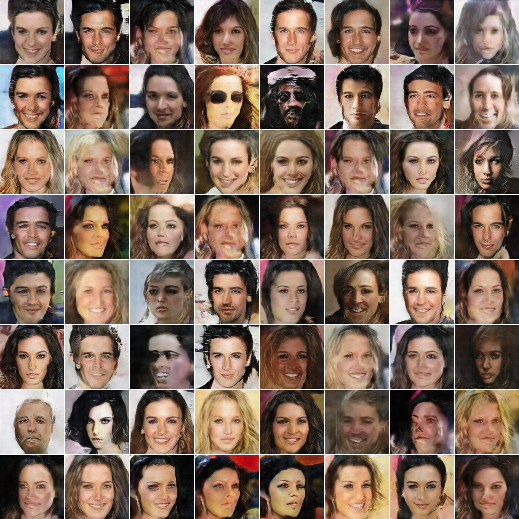} \addhspacesmall
\label{fig:celeba:reconstructions:ourgan}
}\end{subfigure}
\caption{
Training reconstructions obtained using AGE %
and
\ourgan~on CelebA.%
}
\label{fig:celeba:reconstructions}
\end{center}
\end{figure*}

\begin{figure*}[htbp]
\begin{center}
\begin{subfigure}[AGE]{
\includegraphics[width=0.45\textwidth]{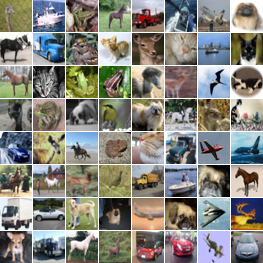} \addhspacesmall
\includegraphics[width=0.45\textwidth]{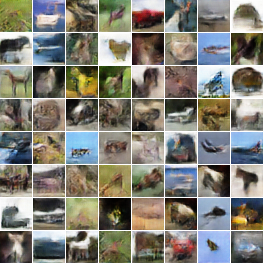} \addhspacesmall
\label{fig:cifar:reconstructions:age}
}\end{subfigure}
\begin{subfigure}[\ourgan]{
\includegraphics[width=0.45\textwidth]{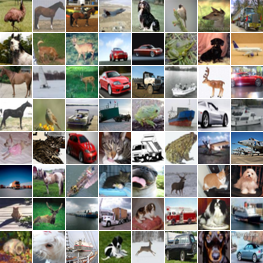} \addhspacesmall
\includegraphics[width=0.45\textwidth]{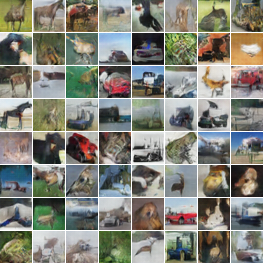} \addhspacesmall
\label{fig:cifar:reconstructions:ourgan}
}\end{subfigure}
\caption{
Training reconstructions obtained using AGE %
and
\ourgan~on CIFAR-10. %
Left is the data and right are reconstructions.
}
\label{fig:cifar:reconstructions}
\end{center}
\end{figure*}

We show reconstructions obtained using \ourgan~and AGE for the CelebA dataset in Figure \ref{fig:celeba:reconstructions} and on CIFAR-10 in Figure \ref{fig:cifar:reconstructions}.

\section{Ablation experiment: code discriminator and the empirical KL\label{app:code_empirical_kl}}

We have shown that we can estimate the KL term in \eqref{eq:free_energy} using the density ratio trick. In the case of a normal prior, another way to estimate the KL divergence on a mini-batch of latents each of dimension $n$, with per dimension sample mean and variance denoted by $m_i$ and $s_i$ ($i = 1 ... n$) respectively, is\footnote{This approximation was also used by \citet{age} in AGE.}:

\begin{align}
\textrm{KL}(q(z|x), N(0, I)) \approx \frac{n}{2} + \sum_{i=1}^{n} (\frac{(s_i)^2 + (m_i)^2}{2} - \log(s_i))
\label{eq:kl_approx}
\end{align}

\begin{figure*}[ht]
\begin{center}
\includegraphics[width=0.45\textwidth]{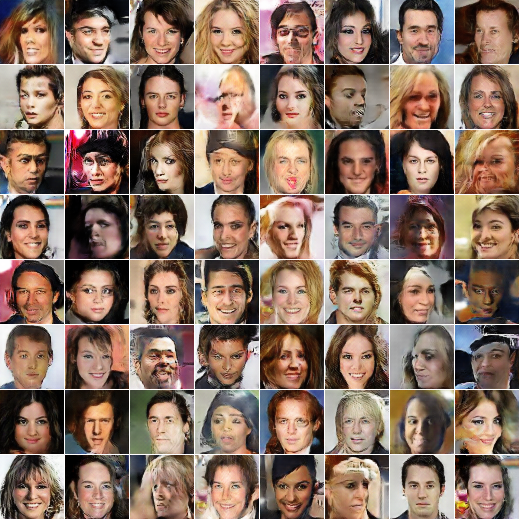} \addhspacesmall
\includegraphics[width=0.45\textwidth]{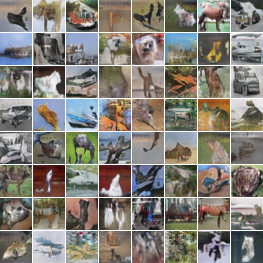} \addhspacesmall
\caption{
Samples from \ourgan~on CelebA and CIFAR-10, trained using the empirical KL approximation (as opposed to a code discriminator) to make the posterior and the prior of the latents match.}
\label{fig:empirical_kl_samples}
\end{center}
\end{figure*}

\begin{figure*}[htbp]
\begin{center}
\includegraphics[width=0.48\textwidth]{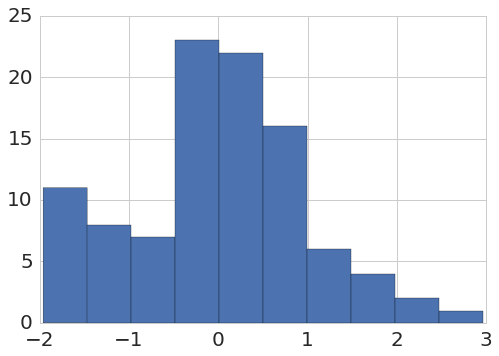} \addhspacesmall
\includegraphics[width=0.48\textwidth]{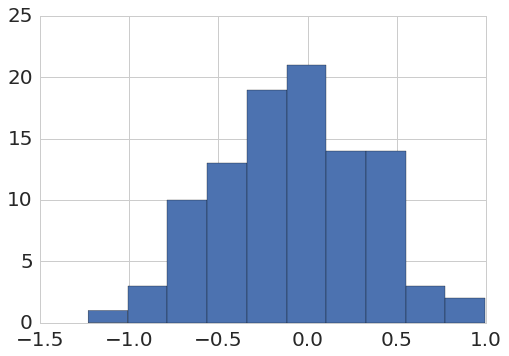} \addhspacesmall
\caption{
 Histogram of latent means obtained on 64000 code representations from  \ourgan~trained using a code discriminator (left) and the empirical KL approximation (right). The latent size was 100. Since the prior was set to a normal with mean 0, we expect most means to be around 0. We note that the empirical KL seems better at forcing the means to be around 0.
}
\label{fig:means:empirical_vs_code}
\end{center}
\end{figure*}

\begin{figure*}[htbp]
\begin{center}
\includegraphics[width=0.48\textwidth]{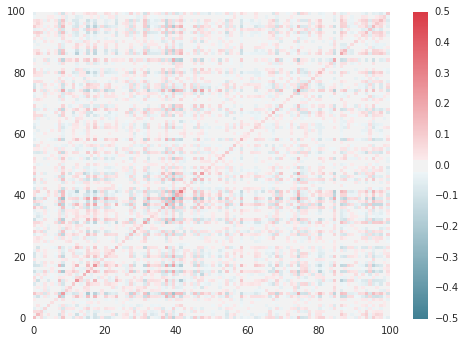} \addhspacesmall
\includegraphics[width=0.48\textwidth]{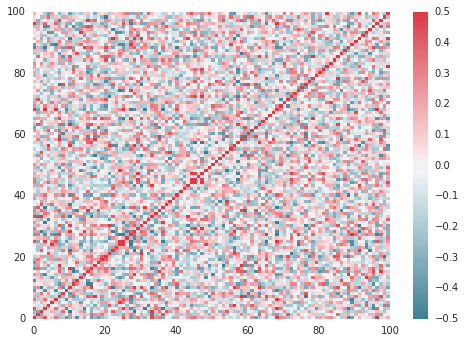} \addhspacesmall
\caption{
 Covariance matrices obtained on 64000 code representations from  \ourgan~trained using a code discriminator (left) and the empirical KL approximation (right). The latent size was 100. We note that the code discriminator produces latents which have a lot less correlation than the empirical KL (which is what we want in this case, since the prior was a univariate Gaussian).
}
\label{fig:cov:empirical_vs_code}
\end{center}
\end{figure*}

In order to understand how the two different ways of estimating the KL term compare, we replaced the code discriminator in \ourgan~with the KL approximation in \eqref{eq:kl_approx}. We then compared the results both by visual inspection (see CelebA and CIFAR-10 samples in Figure \ref{fig:empirical_kl_samples}) and by evaluating how well the prior was matched. In order to avoid be able to use the same hyperparameters for different latent sizes, we divide the approximation in \eqref{eq:kl_approx} by the latent size. To also understand the effects of the two methods on the resulting autoencoder codes, we plot the means (Figure \ref{fig:means:empirical_vs_code}) and the covariance matrix (Figure \ref{fig:cov:empirical_vs_code}) obtained from the a set of saved latent codes. By assessing the statistics of the final codes obtained by models trained using both approaches, we see that the two models of enforcing the prior have different side effects: the latent codes obtained using the code discriminator are decorrelated, while the ones obtained using the empirical KL are entangled; this is expected, since the correlation of latent dimensions is not modeled by \eqref{eq:kl_approx}, while the code discriminator can pick up that highly correlated codes are not from the same distribution as the prior. While the code discriminator achieves better disentangling, the means obtained using the empirical KL are closer to 0, the mean of the prior distribution for each latent. We leave investigating these affects and combining the two approaches for future work.

\section{Monitoring overfitting of the independent Wasserstein critic\label{app:improved_critic}}
\begin{figure*}[htbp]
\begin{center}
\includegraphics[width=0.45\textwidth]{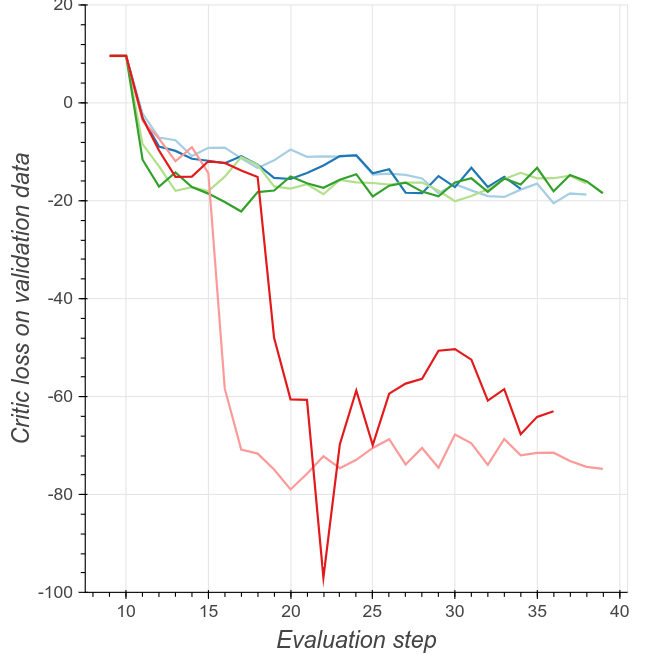} \addhspacesmall
\includegraphics[width=0.45\textwidth]{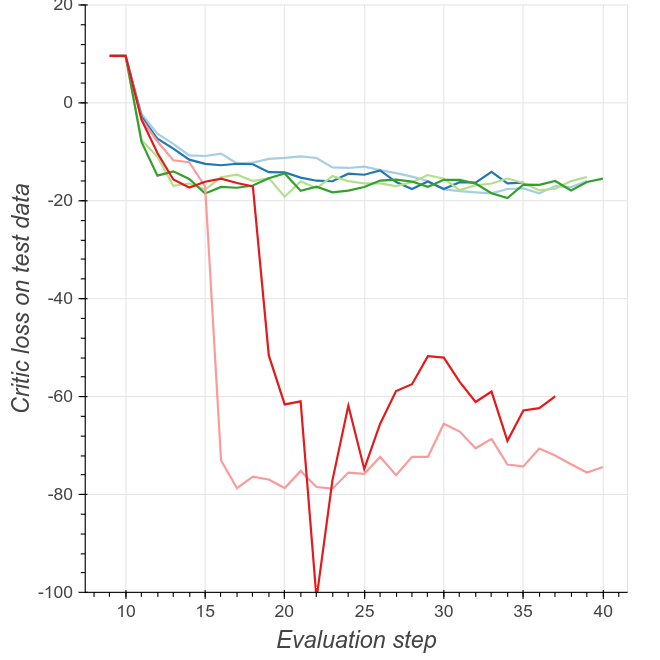} \addhspacesmall
\caption{
 Training curves of the independent Wasserstein critic for different hyperparameter values. The model trained here is \ourgan~, trained on CelebA. Left: the loss obtained on a mini-batch from the validation data. Right: the average loss obtained on the entire test set.
}
\label{fig:wgan_critic_curves}
\end{center}
\end{figure*}

To ensure that the independent Wasserstein critic does overfit during training to the validation data, we monitor the difference in performance between training and test (see Figure ~\ref{fig:wgan_critic_curves}).

\section{Best samples according to different metrics\label{app:sample_metric_compare}}
\begin{figure*}[ht]
\begin{center}
\includegraphics[width=0.31\textwidth]{celeba_our_latest} \addhspace
\includegraphics[width=0.31\textwidth]{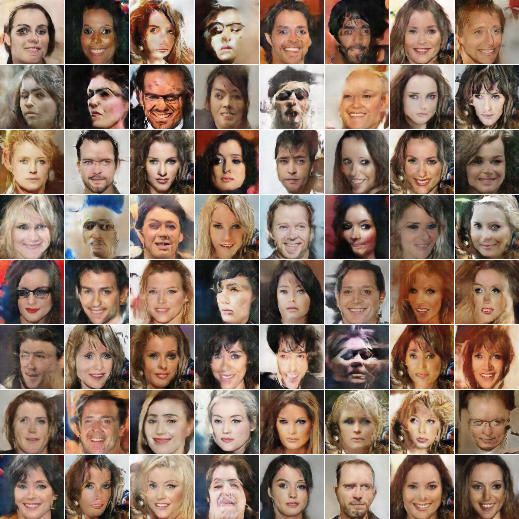} \addhspace
\includegraphics[width=0.31\textwidth]{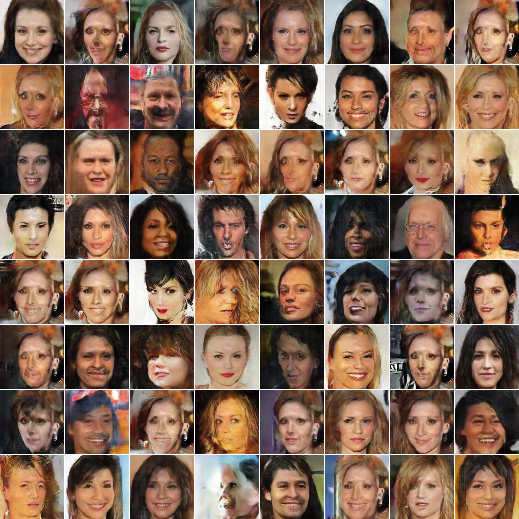} %
\caption{Best samples from \ourgan~trained on CelebA according to different metrics: sample quality (left), independent Wasserstein critic (middle), sample diversity (right) given by 1-MSSSIM.}
\label{fig:celeba:samples:metric}
\end{center}
\end{figure*}

Figure~\ref{fig:celeba:samples:metric} shows the best samples on CelebA according to different metrics.

\section{Relationships between different metrics\label{app:wgan_mssim}}

\begin{figure*}[ht]
\begin{center}
\includegraphics[width=0.5\textwidth]{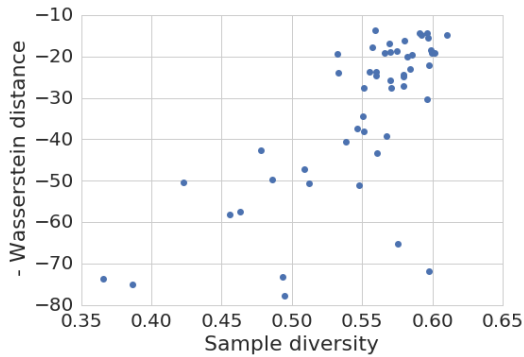}
\caption{Correlation between sample diversity and the negative Wasserstein distance, obtained from a \ourgan~hyperparameter sweep.
}
\label{fig:msssim_wgan}
\end{center}
\end{figure*}

We assess the correlation between sample quality and how good a model is according to a independent Wasserstein critic in Figure~\ref{fig:msssim_wgan}.

\section{Training details: hyperparameters and network architectures\label{app:hyperparameters}}

 For all our models, we kept a fixed learning rate throughout training. We note the difference with AGE, where the authors decayed the learning rate during training, and changed the loss coefficients during training\footnote{As per advice found here: \url{https://github.com/DmitryUlyanov/AGE/}}.). The exact learning rate sweeps are defined in Table ~\ref{tab:learning_rates}. We used the Adam optimizer \citep{adam} with $\beta_1 = 0.5$ and $\beta_2 = 0.9$ and a batch size of 64 for all our experiments. We used batch normalization \citep{batchnorm} for all our experiments. We trained all \colormnist~ models for 100000 iterations, and CelebA and CIFAR-10 models for 200000 iterations.

\begin{table}[h]
\centering
\resizebox{\columnwidth}{!}{
\begin{tabular}{@{}l|c|c|c|c@{}}
  \multicolumn{5}{c}{Model} \\ \midrule
Network  & DCGAN &  WGAN-GP      & \ourgan & AGE \\ \midrule
Generator/Encoder &  $0.0001, 0.0002, 0.0003$ & $0.0001, 0.0002, 0.0003$ &  $0.0001, 0.0005$ & $0.0001, 0.0002, 0.0005$  \\
Discriminator      &  $0.0001, 0.0002, 0.0003$ & $0.0001, 0.0002, 0.0003$ &  $0.0005$ &   \\
Code discriminator  &   &  & $ 0.0005$  &   \\
\end{tabular}
}
\vspace{0.2cm}
\caption{\label{tab:learning_rates} Learning rate sweeps performed for each model.}
\end{table}

\subsection{Scaling coefficients}

We used the following sweeps for the models which have combined losses with different coefficients (for all our baselines, we took the sweep ranges from the original papers):
\begin{itemize}
\item WGAN-GP
  \begin{itemize}
    \item The gradient penalty of the discriminator loss function: 10.
  \end{itemize}

\item AGE
  \begin{itemize}
    \item Data reconstruction loss for the encoder: sweep over 100, 500, 1000, 2000.
    \item Code reconstruction loss for the generator: 10.
  \end{itemize}

\item \ourgan
  \begin{itemize}
    \item Data reconstruction loss for the encoder: sweep over 1, 5, 10, 50.
    \item Data reconstruction loss for the generator: sweep over 1, 5, 10, 50.
    \item Adversarial loss for the generator (coming from the data discriminator): 1.0.
    \item Adversarial loss for the encoder (coming from the code discriminator): 1.0.
  \end{itemize}
\end{itemize}

\subsection{Choice of loss functions}
  For AGE, we used the $l_1$ loss as the data reconstruction loss, and we used the cosine distance for the code reconstruction loss. For \ourgan~, we used $l_1$ as the data reconstruction loss and the traditional GAN loss for the data and code discriminator.

\subsection{Choice of latent prior}
  We use a normal prior for all models, apart from AGE \citep{age} which uses a uniform unit ball as the prior, and thus we project the output of the encoder to the unit ball.

\subsection{Network architectures}

For all our baselines, we used the same discriminator and generator architectures, and we controlled the number of latents for a fair comparison. For AGE we used the encoder architecture suggested by the authors\footnote{Code at: \url{https://github.com/DmitryUlyanov/AGE/}}, which is very similar to the DCGAN discriminator architecture. For \ourgan~, the encoder is always set as a convolutional network, formed by transposing the generator (we do not use any activation function after the encoder). All discriminators and the AGE encoder use leaky units with a slope of 0.2, and all generators used ReLUs. For all our experiments using \ourgan~, we used as a code discriminator a 3 layer MLP, each layer containing 750 hidden units. We did not tune the size of this network, and we postulate that since the prior latent distributions are similar (multi variate normals) between datasets, the impact of the architecture of the code discriminator is of less importance than the architecture of the data discriminator, which has to change from dataset to dataset (with the complexity of the data distribution). However, one could improve on our results by carefully tuning this architecture too.

\subsubsection{\colormnist~}

For all our models trained on \colormnist, we swept over the latent sizes 10, 50 and 75. Tables~\ref{tab:color_mnist_disc} and \ref{tab:color_mnist_gen} describe the discriminator and generator architectures respectively.

\begin{table}[h]
\centering
\begin{tabular}{@{}rlll@{}} \toprule
Operation              & Kernel       & Strides      & Feature maps  \\ \midrule
Convolution            & $5 \times 5$ & $2 \times 2$ & $8$   \\
Convolution            & $5 \times 5$ & $1 \times 1$ & $16$   \\
Convolution            & $5 \times 5$ & $2 \times 2$ & $32$    \\
Convolution            & $5 \times 5$ & $1 \times 1$ & $64$    \\
Convolution            & $5 \times 5$ & $2 \times 2$ & $64$    \\
Linear adv            & N/A          & N/A          & $2$     \\
Linear class     & N/A          & N/A          & $10$         \\
\end{tabular}
\vspace{0.2cm}
\caption{\label{tab:color_mnist_disc} \colormnist~discriminator architecture used for DCGAN,  WGAN-GP and \ourgan. For DCGAN, we use dropout of 0.8 after the last convolutional layer. No other model uses dropout.}
\end{table}

\begin{table}[h]
\centering
\begin{tabular}{@{}rlll@{}} \toprule
Operation              & Kernel       & Strides      & Feature maps    \\ \midrule
Linear                 & N/A          & N/A          & $3136$       \\
Transposed Convolution & $5 \times 5$ & $2 \times 2$ & $64$          \\
Transposed Convolution & $5 \times 5$ & $1 \times 1$ & $32$          \\
Transposed Convolution & $5 \times 5$ & $2 \times 2$ & $3$           \\
\end{tabular}
\vspace{0.2cm}
\caption{\label{tab:color_mnist_gen} \colormnist~generator architecture. This architecture was used for all 4 compared models.}

\end{table}

\subsubsection{CelebA and CIFAR-10}

The discriminator and generator architectures used for CelebA and CIFAR-10 were the same as the ones used by \citet{improvedwgan} for WGAN.\footnote{Code at: \url{https://github.com/martinarjovsky/WassersteinGAN/blob/master/models/dcgan.py}}. Note that the WGAN-GP paper reports Inception Scores computed on a different architecture, using 101-Resnet blocks.

\subsection{CIFAR-10 classifier used for Inception score \label{app:cifar_10_net}}

We used a VGG style \citep{vgg} convnet trained on CIFAR-10 as the classifier network used to report the inception score in Section~\ref{sec:metrics}. The architecture is described in Table~\ref{tab:cifar_classifier_net}. We use batch normalization after each convolutional layer. The data is rescaled to be in range $[-1, 1]$, and during training the input images are randomly cropped to size $(24, 24, 3)$. We used a momentum optimizer with learning rate starting at 0.1 and decaying by 0.1 at timesteps 40000 and 60000, with momentum set at 0.9. We used an $l_2$ regularization penalty of $1e-4$. The network was trained for 80000 epochs, using a batch size of 256 (8 synchronous workers, each having a batch size of 32). The resulting network achieves an accuracy of 5.5\% on the official CIFAR-10 test set.

\begin{table}[h]
\centering
\begin{tabular}{@{}rlll@{}} \toprule
Operation              & Kernel       & Strides      & Feature maps  \\ \midrule
Convolution            & $3 \times 3$ & $2 \times 2$ & $64$   \\
Convolution            & $3 \times 3$ & $1 \times 1$ & $64$   \\
Convolution            & $3 \times 3$ & $2 \times 2$ & $128$    \\
Convolution            & $3 \times 3$ & $1 \times 1$ & $128$    \\
Convolution            & $3 \times 3$ & $2 \times 2$ & $128$    \\
Convolution            & $3 \times 3$ & $2 \times 2$ & $256$    \\
Convolution            & $3 \times 3$ & $2 \times 2$ & $256$    \\
Convolution            & $3 \times 3$ & $2 \times 2$ & $256$    \\
Convolution            & $3 \times 3$ & $2 \times 2$ & $512$    \\
Convolution            & $3 \times 3$ & $2 \times 2$ & $512$    \\
Convolution            & $3 \times 3$ & $2 \times 2$ & $512$    \\

Average pooling        & N/A          & N/A          & N/A        \\
Linear class          & N/A          & N/A          & $10$         \\
\end{tabular}
\vspace{0.2cm}
\caption{\label{tab:cifar_classifier_net} The neural network trained to classify CIFAR-10 data. }
\end{table}

\end{document}